%% file: neurips_2026.tex
\documentclass{article}

\usepackage[preprint]{neurips_2026}

\usepackage[utf8]{inputenc} 
\usepackage[T1]{fontenc}    
\usepackage{hyperref}       
\usepackage{url}            
\usepackage{booktabs}       
\usepackage{amsfonts}       
\usepackage{nicefrac}       
\usepackage{microtype}      
\usepackage{xcolor}         
\usepackage{graphicx}
\usepackage{array}
\usepackage{caption}
\usepackage{enumitem}
\setlist{itemsep=1pt, topsep=2pt, parsep=0pt, partopsep=0pt}

\title{Mirage Probes: How Vision Models Fake\\ Visual Understanding}

\author{%
  Daniel Ben-Levi\textsuperscript{1}\thanks{Correspondence to db3651@columbia.edu.} \And
  Judah Goldfeder\textsuperscript{1} \And
  Weiliang Zhao\textsuperscript{1} \And
  Raz Lapid\textsuperscript{2} \And
  Amit LeVi\textsuperscript{3} \And
  Allen G Roush\textsuperscript{4} \And
  Ravid Shwartz-Ziv\textsuperscript{5} \And
  Hod Lipson\textsuperscript{1} \And
  {\normalfont
  \textsuperscript{1}Columbia University \quad
  \textsuperscript{2}Intuit \quad
  \textsuperscript{3}Technion \quad
  \textsuperscript{4}Thoughtworks \quad
  \textsuperscript{5}New York University} \\
}

\begin{document}

\maketitle

\begin{abstract}
Vision-language models (VLMs) can answer image-based questions confidently, and often correctly, even when no image is provided. This mirage behavior inflates benchmark scores without reflecting visual grounding. Prior work treats this as a single failure mode. We argue it is two. Using \emph{Mirage Probes}, a contrastive probing framework that pairs paraphrased question variants with matched mirage and non-mirage labels on the same image, we show that mirage behavior is linearly decodable from internal activations across residual stream, MLP, post-attention, and attention-head sites in two open-source VLMs. We demonstrate that a Naive Bayes text baseline cannot recover this signal, ruling out surface lexical confounds. Cross-benchmark separability patterns, together with a novel Prior Harnessing Index (PHI) measuring how much a model can answer from text alone, expose two distinct regimes: textual biases, where the model answers from language priors without engaging visual representations, and spurious images, where it constructs false visual content in latent space and answers as if grounded. The distinction has direct mitigation consequences: text-distribution cleaning can address the first regime but cannot reach the second, since spurious-image mirages live in the model's visual representations rather than its text. Faithful visual grounding will require interventions at the representational level. Code is available \href{https://github.com/danielbenlevi/mirage_probes}{here}.
\end{abstract}

\input{sections/intro}
\input{sections/related_work}

\input{sections/method}

\input{sections/results}

\input{sections/conclusions}

\input{sections/limitations}

\bibliographystyle{plainnat}
\bibliography{neurips_2026}


\appendix
\newpage
\appendix
\section*{Appendix}
\input{sections/appendix}





\end{document}

%% file: sections/intro.tex
\section{Introduction}

Vision-language models (VLMs) are increasingly used for tasks that require faithful visual understanding, including visual question answering \citep{antol2015vqa,kafle2017visual}, document analysis \citep{mathew2021docvqa,kim2022ocr}, scientific reasoning \citep{lu2022learn}, and medical image interpretation \citep{li2023llava,moor2023med,zhang2023pmc}. In such settings, a correct answer is insufficient unless it is grounded in the visual input. Yet recent work on mirage behavior \citep{asadi2026mirage} challenges the assumption that benchmark success reflects faithful visual reasoning: VLMs can answer image-based questions confidently, and sometimes correctly, even when the image is absent. This suggests that performance on multimodal benchmarks such as MMMU \citep{yue2024mmmu} and VQA-RAD \citep{lau2018dataset} may partly reflect textual priors, dataset regularities, or benchmark-specific shortcuts rather than genuine use of visual evidence.

\begin{figure}[t]
    \centering
    \includegraphics[width=.75\linewidth]{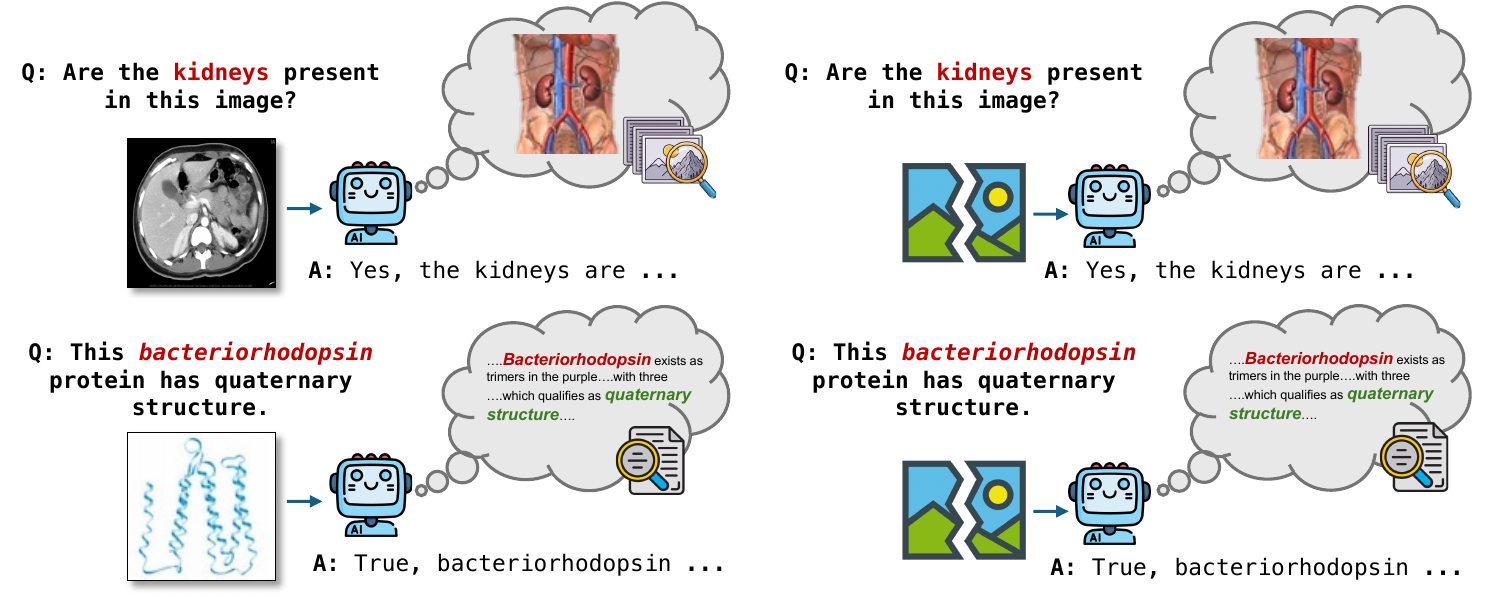}
    \caption{\textbf{Two distinct mirage mechanisms.} VLMs seem to exhibit two different kinds of mirage behavior, \textit{spurious images} and \textit{textual biases}, depicted here. Each row shows a mirage response to the same question with and without an image present. In the first example, the model achieves mirage behavior by building and referring to a false visual representation. Differently, in the second example, the model does not utilize visual information at all, relying solely on textual priors provided by the rich question distribution to reach the correct response.}
    \label{fig:example2}
    \vspace{-3mm}
\end{figure}

To remediate this behavior, we must understand what gives rise to it internally. Prior work largely treats mirages as a single failure mode \citep{asadi2026mirage}, but the same output-level behavior may arise from different mechanisms. A model may form internal representations consistent with unsupported visual content, effectively behaving as though a \emph{spurious image} were present. Alternatively, it may answer directly from \emph{textual biases} (benchmark-specific or distributional priors) without forming any false visual representation. These mechanisms, illustrated in Figure~\ref{fig:example2}, imply different interventions: text-distribution cleaning may reduce shortcut-driven mirages, but may be insufficient when mirage behavior is associated with unsupported visual inference.

In this work, we introduce \emph{Mirage Probes}, a representation-level framework for studying mirage behavior in VLMs. We construct datasets of mirage-associated and non-mirage-associated image-conditioned responses by comparing model generations with and without visual input, and we build contrastive pairs from lightly paraphrased question variants that preserve image-question semantics while yielding different mirage labels, reducing surface-level textual confounds. We then probe internal activations from open-source VLMs at four sites (namely, residual stream states, MLP outputs, post-attention outputs, and individual attention-head outputs), using probe families that test whether mirage information is encoded as a single linear direction, requires a nonlinear transformation, or is most cleanly recovered from the relational structure between matched contrastive examples.

Our results show that mirage behavior is linearly decodable from VLM latent representations across all four activation sites when images are provided, indicating that it is not merely an image-absent or output-level artifact. Two-layer MLP probes do not meaningfully outperform linear ones, and contrastive difference probes recover the signal most cleanly. Together, these results are consistent with mirage information being encoded along recoverable linear directions rather than requiring nonlinear extraction. A Naive Bayes textual baseline trained on response text is consistently weaker than our contrastive probes, suggesting that this signal is not reducible to surface lexical features. Separability nonetheless varies sharply across benchmarks: it is strongest on VQA-RAD \citep{lau2018dataset}, where questions typically require specific visual evidence and provide little textual context, and weaker on benchmarks where answers can more often be inferred from textual priors. We propose, as a hypothesis motivating further mechanistic work, that behavioral mirages conflate at least two representational regimes -- spurious images and textual biases -- rather than constituting a single phenomenon. More broadly, our findings raise concerns for VLM training and evaluation: reducing textual shortcuts is necessary, but may not be sufficient to ensure faithful visual grounding.

Our contributions are:
\begin{enumerate}
    \item \textbf{We show mirage behavior is linearly decodable from image-present VLM latent space.}
    Linear probes distinguish mirage-associated from non-mirage-associated image-conditioned generations across multiple activation sites, and contrastive difference probes recover this signal most cleanly.

    \item \textbf{We introduce \emph{Mirage Probes}, a contrastive probing framework for mirage behavior.}
    By pairing semantically similar question variants with different mirage labels, our framework reduces surface-level textual confounds, as confirmed by comparison against a Naive Bayes textual baseline.

    \item \textbf{We hypothesize two distinct forms of mirage behavior.}
    Cross-benchmark separability patterns suggest that prior behavioral definitions group together \emph{spurious-image} mirages and \emph{textual-bias} mirages, though we leave causal validation of this distinction to future work.

    \item \textbf{We argue that text-only mitigation may be mechanism-incomplete.}
    Text-distribution cleaning targets textual-bias mirages but is unlikely to address spurious-image mirages, since the latter are associated with internal representations rather than surface textual patterns.
\end{enumerate}

%% file: sections/related_work.tex
\section{Related work}

\paragraph{Hallucination in vision-language models.}

Hallucination in VLMs has been studied extensively, primarily as the assertion of visual content not present in the input image. Object hallucination benchmarks such as POPE \citep{li2023evaluating} and CHAIR \citep{rohrbach2018object} probe whether models invent objects, while subsequent work has extended the analysis to attribute and relational hallucinations \citep{sun2024aligning, jiang2024hallucination}. Recent works provide broad treatments of the phenomenon, its causes, and mitigations \citep{liu2024survey, bai2024hallucination, goldfeder2026ai}. 

\paragraph{Image-absent generation and modality bypass.}

A growing line of work examines cases in which VLMs appear to bypass visual input. \citet{tong2024eyes} show that state-of-the-art VLMs frequently fail on visual questions that humans answer trivially. Most directly, \citet{asadi2026mirage} introduce the term \emph{mirage} for the observation that VLMs produce confident, often correct answers to image-grounded questions even when no image is provided, and hypothesize that the same mechanism may operate when images are present. To our knowledge, no prior work has tested this hypothesis at the level of internal representations.

\paragraph{Probing and mechanistic interpretability of multimodal models.}

Linear probes on intermediate representations have a long history in NLP interpretability, from early diagnostic classifiers \citep{alain2016understanding, belinkov2022probing} to syntactic probes \citep{hewitt2019structural} and the broader study of what linear directions encode in language model representations \citep{tenney2019bert, park2023linear,lomasov2025exploring, theodoridis2026probing}. Contrastive probe constructions \citep{burns2024discoveringlatentknowledgelanguage, marks2024geometry} have proven particularly effective at recovering features otherwise masked by surface confounds, and we adopt a contrastive-pair design for the same reason.  We extend this line of work by targeting a specific behavioral failure, mirage generation, and showing that it has a recoverable latent representation that further admits a two-mechanism decomposition.

%% file: sections/method.tex
\section{Methods}
\label{sec:methods}

\subsection{Mirage Behavior}

Let $M$ denote a vision-language model. For each example $i$, let $x_i$ be the image, $q_i$ be the textual question, and let
\[
r_i^{\mathrm{img}} = M(x_i, q_i), \qquad
r_i^{\emptyset} = M(\emptyset, q_i)
\]
denote the model response with and without the image, respectively. Mirage behavior occurs when the model produces an image-grounded answer even when the image is absent. In this case, the model may appear to use visual evidence, while actually relying on textual priors, dataset regularities, or unsupported internal visual assumptions.

We define a binary mirage label $y_i \in \{0,1\}$ for each with-image response. The label $y_i=1$ denotes a mirage-like response, and $y_i=0$ denotes a non-mirage-like response. Since a with-image mirage cannot be directly observed from the output alone, this label is treated as a behavioral proxy obtained by comparing $r_i^{\mathrm{img}}$ and $r_i^{\emptyset}$. The goal of our probing setup is to test whether this behavioral label is recoverable from the model's internal representations.

\subsection{Dataset Construction}
\label{subsec:dataset_construction}

\paragraph{Annotation Scheme.} Our goal is to assemble labeled examples of mirage and non-mirage with-image responses. Whether a given with-image generation is actually a mirage cannot be determined from the output alone, so we rely on a heuristic. Following the hypothesis from \citet{asadi2026mirage} that distribution shift between with-image and without-image responses signals genuine reliance on provided visual input, we adopt the following labeling scheme.

A response is \textbf{mirage-like} if the with-image and without-image generations produce the same answer and have a bag-of-words vector cosine similarity above 0.7. A response is \textbf{non-mirage-like} if the without-image generation explicitly expresses uncertainty due to missing visual context or states that answering is impossible without the image, both detected through regex checks. A response is \textbf{ambiguous} if the without-image generation is itself mirage-like but the with-image answer differs in either answer or content similarity. We choose to include an explicit ambiguous label because in such cases, it is impossible to determine whether the with-image generation represents a legitimate use of visual information or is just a different mirage answer triggered by the distribution shift of including the image.

We also tried two alternative schemes. Folding ambiguous responses into the mirage-like class and labeling non-mirage examples with the B-Clean criterion described by \citet{asadi2026mirage} (questions that cannot be answered correctly without the image are considered safe), both reduced probe accuracy substantially. Thus, we select the scheme above for our main experiments.

\paragraph{Benchmarks.} We draw questions from four widely-used visual question answering benchmarks chosen to span medical and general expertise, along with free-form and multiple-choice formats. \textbf{VQA-RAD} \citep{lau2018dataset} contains clinically generated free-form and yes/no questions about radiology images. \textbf{MMMU-Pro} \citep{yue2025mmmupro} is a multiple-choice expert-level reasoning benchmark spanning college-level subjects across many disciplines, designed to be more robust to text-only shortcut solutions than its predecessor \citep{yue2024mmmu}. \textbf{MedXpertQA} \citep{zuo2025medxpertqa} provides expert-level multiple-choice medical questions targeted at reasoning rather than recall. \textbf{MicroVQA} \citep{burgess2025microvqa} consists of multiple-choice questions over biomedical microscopy images.

\begin{figure}[t]
    \centering
    \includegraphics[width=0.85\linewidth]{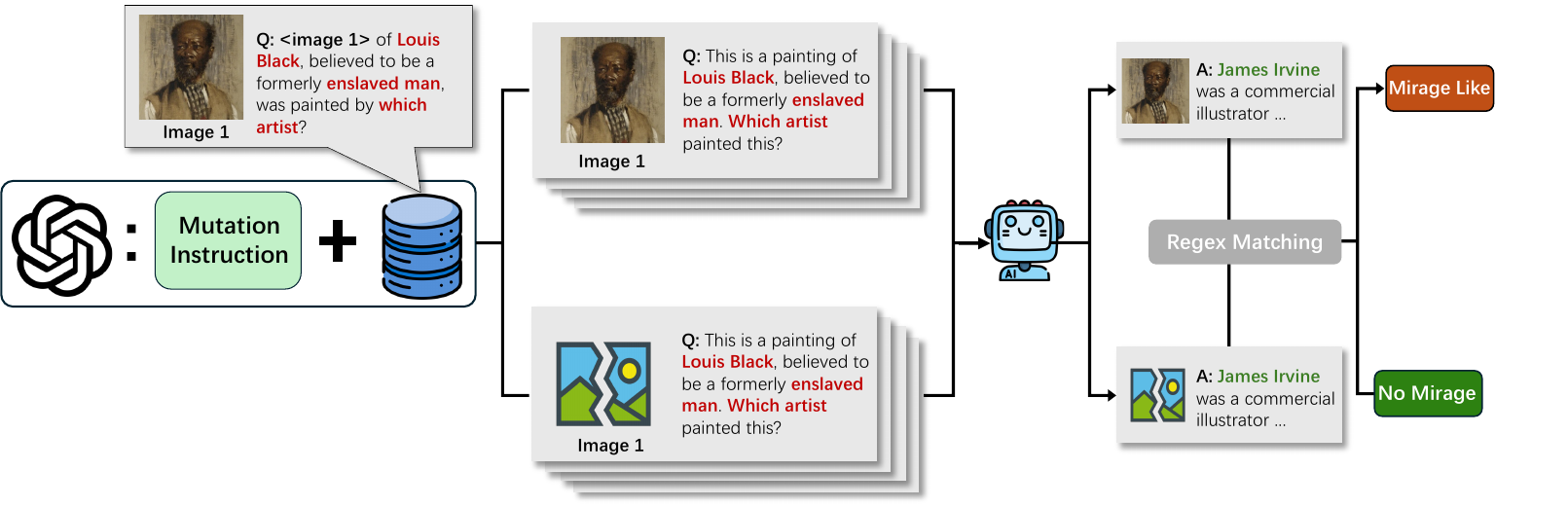}
    \caption{\textbf{Contrastive dataset construction overview.} In order to produce contrastive pairs, we mutate each base question four times with GPT-4o-mini and generate with and without-image responses to each variant. Then, for each pair of responses, we use cosine similarity and a regex annotator to identify mirage and non-mirage behavior. If a base question's group contains at least one mirage and non-mirage pair, we store each pair's with-image response to form a contrastive example.}
    \label{fig:dataset2}
\end{figure}

\paragraph{Contrastive Pairs.} Linear probes benefit substantially from training on contrastive pairs, where positive and negative examples differ only in the property of interest \citep{burns2024discoveringlatentknowledgelanguage, marks2024geometry}. Without contrastive structure, surface-level features (topic words, response length, question style) often correlate with labels, causing probes to learn shortcuts rather than the target representation.

For each base question, we generate four mutations using GPT-4o-mini \citep{openai2024gpt4ocard}. A mutation rephrases or rearranges the question while preserving its meaning and adding no new information. The intent is mild distributional perturbation, sufficient to flip behavior between mirage and non-mirage for some questions while keeping the image and the underlying intent fixed. When the responses to a base question and its four mutations together contain at least one mirage-like and one non-mirage-like response, we draw a single contrastive pair from that question set. We never draw more than one pair per base question to preserve diversity.

We construct two datasets per evaluated model. The \textbf{contrastive-pairs} dataset contains only such pairs, with the same image and near-identical question content across both classes. The \textbf{all-examples} dataset contains every response annotated as mirage-like or non-mirage-like, capped at one example per class per base question. The contrastive split aims to test whether the mirage signal survives without surface-level confounds, while the all-examples split contains a larger quantity of examples, which may aid generalization if textual confounds do not override the desired direction. Our dataset construction procedure is summarized in Figure~\ref{fig:dataset2}.

\subsection{Model Architecture and Activation Extraction}
\label{subsec:model_architecture}

We probe three open-source vision-language models spanning different model sizes and architectures. \textbf{Ovis2.5-2B} \citep{lu2024ovis} is a 2B-parameter VLM that aligns vision and language representations through a structured embedding scheme. \textbf{GLM-4.6-flash} \citep{glm2024chatglmfamilylargelanguage} is the smaller variant in the GLM multimodal series. \textbf{Qwen3-32B-VL-Instruct} \citep{bai2025qwen3vltechnicalreport} is the 32B-scale vision-language model in the Qwen3 family. Standard VLM design glues a vision encoder to a language model through a learned projection that maps image patch features into the language model's embedding space \citep{li2022blip, liu2023visual}. In our experiments, image tokens are prepended to the user's textual query, after which the language model performs standard causal attention.

We extract activations at four targets within each model: the \textbf{residual stream} at every layer, the \textbf{MLP output} at every layer, the \textbf{post-attention output} at every layer, and \textbf{every individual attention head output} at every layer. We initially included the vision encoder and the projection layers but found their representations carried no detectable mirage signal under any probing strategy and accordingly dropped them from our main analysis.

For each model and each annotated dataset, we run the model on chat-formatted conversations and cache activations under three pooled token-position aggregations, building on the position scheme introduced by \citet{theodoridis2026probing}. The first aggregation, \texttt{text-nonspecial-mean}, averages activations over all non-special response text tokens. The second, \texttt{vision-tail}, averages over all vision tokens and the last special token in the response. The third, \texttt{vision-text-mean}, averages over all vision tokens and all non-special response text tokens jointly. Our activation extraction procedure is illustrated in Figure~\ref{fig:probe2}.

Final per-model dataset sizes are presented in Appendix~\ref{app:dataset_sizes}. GLM-4.6-flash produces too few non-mirage responses on three of four benchmarks to support stable probing, so we drop it from the main results. MicroVQA yields too few non-mirage responses for stable probing across models, so it is also dropped from the main results. Qualitatively, we note that MicroVQA has the lowest visual reliance among the four benchmarks, which is likely related to its extremely high associated mirage rate. Ovis2.5-2B produces a relatively small viable dataset for some benchmarks, but we retain the model nonetheless, as consistent results across Ovis2.5-2B and Qwen3-32B-VL-Instruct rule out scale-specific artifacts and increase the robustness of our findings.

\begin{figure}[t]
    \centering
    \includegraphics[width=0.95\linewidth]{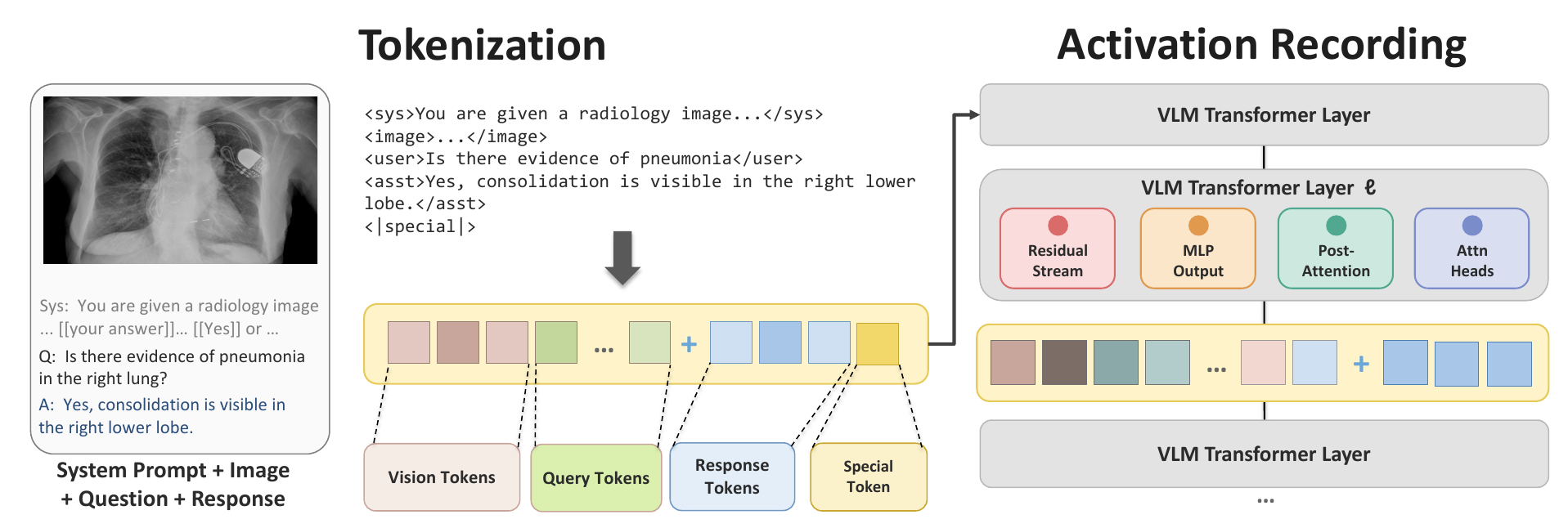}
    \caption{\textbf{Activation extraction overview.} The VLM first generates a response to a user query, following system prompt instructions and referring to attached images if any are provided. The full conversation, model response included, is then chat-template tokenized and fed into the VLM for extraction. Any images are first processed by the model's vision encoder and then projected into vision tokens, which are prepended to the user's query. Then, the model conducts an LLM decoder-only pass over the full conversation, and we extract activations from the residual stream, MLP outputs, post-attention outputs, and individual attention head outputs at each layer.}
    \label{fig:probe2}
\end{figure}

\subsection{Latent Space Probes}

We train four families of probes, each motivated by a different prior on how mirage information could be encoded. Let $h_{\ell,i} \in \mathbb{R}^{d_\ell}$ denote the activation of example $i$ extracted from layer or component $\ell$. In our setting, $\ell$ may correspond to a residual stream state, an MLP output, a post-attention output, or an individual attention-head output. A probe is a classifier $g_\phi$ trained to predict the mirage label from the activation:
\[
\hat{y}_i = g_\phi(h_{\ell,i}).
\]
Thus, probe accuracy measures whether mirage behavior is decodable from the model's latent space.

\textbf{Logistic regression} is a standard linear probing strategy \citep{alain2016understanding, goldowsky2025detecting}. It predicts
\[
p(y_i=1 \mid h_{\ell,i}) = \sigma(w^\top h_{\ell,i}+b),
\]
where $w$ and $b$ are learned parameters and $\sigma$ is the sigmoid function. Above-chance accuracy implies the target property is linearly decodable from the model's representation space, which is the cleanest evidence that the model encodes that property as a feature direction.

\textbf{Two-layer MLP} probes test whether mirage information is decodable through a lightweight nonlinear transform but not as a single linear separator. A gap between linear and MLP accuracy points to distributed or nonlinear encoding.

\textbf{Logistic regression on concatenated activations} trains a linear probe on the concatenation of activations from all model layers at once. If the model has $L$ layers, we define
\[
\bar{h}_i = [h_{1,i}; h_{2,i}; \ldots; h_{L,i}],
\]
and train logistic regression on $\bar{h}_i$. Such probes allow us to determine whether mirage behavior is sparsely represented across layers instead of cleanly identifiable at a specific point.

\textbf{Logistic regression on the difference of activations} trains on the elementwise difference between with-image and without-image activations for each example. Let $h_{\ell,i}^{\mathrm{img}}$ denote the activation when the image is provided, and let
$h_{\ell,i}^{\emptyset}$ denote the activation when the image is removed. We define
\[
\Delta h_{\ell,i}
=
h_{\ell,i}^{\mathrm{img}} - h_{\ell,i}^{\emptyset}.
\]
A linear probe trained on $\Delta h_{\ell,i}$ tests whether the representational shift caused by adding the image separates mirage-like responses from non-mirage-like responses.

\begin{table*}[t]
\centering

\caption{\textbf{Residual and attention probing results.} We train four probing strategies on our contrastive and all-examples datasets, filtered to each individual benchmark first and then all at once. Results targeting the residual stream are presented on the left, while results on attention heads and post-attention outputs are shown on the right. Our results indicate that mirage behavior is linearly represented in image-present latent space, with consistent non-trivial accuracy across benchmarks and targets.}
\label{tab:residual-attention-probing-side-by-side}

\begin{minipage}[htbp]{0.475\textwidth}
\centering
\textbf{Residual}\\[1mm]
\begingroup
\fontsize{6.9}{7.8}\selectfont
\setlength{\tabcolsep}{0.8pt}
\renewcommand{\arraystretch}{0.86}
\setlength{\aboverulesep}{0pt}
\setlength{\belowrulesep}{2pt}
\begin{tabular*}{\linewidth}{@{\extracolsep{\fill}}lcccc@{}}
\toprule
Strategy & VQA-RAD & MMMU-Pro & MedXpertQA & All \\
\midrule

\multicolumn{5}{c}{\textbf{Ovis -- Contrastive}} \\
\midrule
LogReg & \textbf{75.00\%} & 65.00\% & 68.00\% & 72.20\% \\
MLP & 72.20\% & 62.60\% & 64.00\% & \textbf{72.80\%} \\
Concat & \textbf{67.00\%} & 58.20\% & 52.00\% & 64.00\% \\
Diff & \textbf{97.40\%} & 90.80\% & 71.00\% & 91.80\% \\

\midrule
\multicolumn{5}{c}{\textbf{Ovis -- All Examples}} \\
\midrule
LogReg & 68.09\% & \textbf{73.10\%} & 57.14\% & 71.36\% \\
MLP & 67.23\% & \textbf{75.17\%} & 60.00\% & 70.12\% \\
Concat & 58.72\% & 64.14\% & 51.43\% & \textbf{68.64\%} \\
Diff & \textbf{98.30\%} & 88.97\% & 57.14\% & 92.10\% \\

\midrule
\multicolumn{5}{c}{\textbf{Qwen -- Contrastive}} \\
\midrule
LogReg & \textbf{72.00\%} & 57.80\% & 57.20\% & 61.20\% \\
MLP & \textbf{72.40\%} & 59.60\% & 57.40\% & 60.40\% \\
Concat & \textbf{67.40\%} & 53.40\% & 50.60\% & 57.00\% \\
Diff & \textbf{96.00\%} & 79.40\% & 76.20\% & 77.40\% \\

\midrule
\multicolumn{5}{c}{\textbf{Qwen -- All Examples}} \\
\midrule
LogReg & 61.43\% & \textbf{65.06\%} & 63.17\% & 59.04\% \\
MLP & 73.80\% & \textbf{76.35\%} & 70.43\% & 72.69\% \\
Concat & 70.14\% & \textbf{72.87\%} & 68.70\% & 70.41\% \\
Diff & \textbf{95.77\%} & 80.70\% & 76.52\% & 84.38\% \\

\bottomrule
\end{tabular*}
\endgroup
\end{minipage}
\hfill
\begin{minipage}[htbp]{0.475\textwidth}
\centering
\textbf{Attention}\\[1mm]
\begingroup
\fontsize{6.9}{7.8}\selectfont
\setlength{\tabcolsep}{0.8pt}
\renewcommand{\arraystretch}{0.86}
\setlength{\aboverulesep}{0pt}
\setlength{\belowrulesep}{2pt}
\begin{tabular*}{\linewidth}{@{\extracolsep{\fill}}lcccc@{}}
\toprule
Strategy & VQA-RAD & MMMU-Pro & MedXpertQA & All \\
\midrule

\multicolumn{5}{c}{\textbf{Ovis -- Contrastive}} \\
\midrule
LogReg & \textbf{74.67\%} & 68.00\% & 70.00\% & 73.67\% \\
MLP & \textbf{77.33\%} & 69.00\% & 73.33\% & 71.00\% \\
Concat & \textbf{75.67\%} & 66.33\% & 66.67\% & 72.33\% \\
Diff & 87.67\% & \textbf{91.33\%} & 71.67\% & 85.00\% \\

\midrule
\multicolumn{5}{c}{\textbf{Ovis -- All Examples}} \\
\midrule
LogReg & 75.18\% & 77.01\% & \textbf{85.71\%} & 74.07\% \\
MLP & 75.18\% & 75.86\% & \textbf{90.48\%} & 73.25\% \\
Concat & 73.76\% & 71.26\% & \textbf{80.95\%} & 73.25\% \\
Diff & \textbf{90.78\%} & 87.36\% & 90.48\% & 88.89\% \\

\midrule
\multicolumn{5}{c}{\textbf{Qwen -- Contrastive}} \\
\midrule
LogReg & \textbf{81.00\%} & 63.67\% & 69.33\% & 64.33\% \\
MLP & \textbf{75.67\%} & 66.67\% & 71.67\% & 64.00\% \\
Concat & \textbf{75.33\%} & 64.33\% & 65.00\% & 61.33\% \\
Diff & \textbf{95.67\%} & 81.67\% & 83.33\% & 82.33\% \\

\midrule
\multicolumn{5}{c}{\textbf{Qwen -- All Examples}} \\
\midrule
LogReg & 69.05\% & \textbf{69.62\%} & 67.07\% & 61.52\% \\
MLP & \textbf{75.12\%} & 75.07\% & 74.88\% & 73.97\% \\
Concat & 73.71\% & \textbf{75.94\%} & 71.01\% & 73.52\% \\
Diff & \textbf{95.77\%} & 82.32\% & 78.74\% & 85.39\% \\

\bottomrule
\end{tabular*}
\endgroup
\end{minipage}
\vspace{2mm}
\end{table*}

\subsection{Experimental Setup}
\label{sec:exp_setup}

We train all four types of latent space probes on per-layer activations generated with Ovis and Qwen on conversations from our contrastive and all-examples datasets. Data points where the with-image response is fewer than 10 tokens long are excluded. For each combination of probing strategy, model, and dataset, we conduct four training runs, one on examples exclusively from each of our three retained benchmarks and one on all three benchmarks combined. MLP probes are trained with 2 layers, a hidden dimension of 512, and a GELU activation function. All probes are trained with an Adam optimizer and a learning rate of 0.03, and features are train-set normalized before training.

\textbf{Residual stream probes.} When targeting the residual stream, we aggregate activations with all three options (\texttt{text-nonspecial-mean}, \texttt{vision-tail}, \texttt{vision-text-mean}). Contrastive probes are trained using a 90/10 train/valid split (approximated with hard-coded values for Ovis) and then evaluated on a held-out class-balanced, benchmark-stratified dataset that consist of up to 100 data points from the all-examples dataset. This procedure allows us to maintain robust evaluation while also using our limited contrastive data to the greatest effect. All-examples probes are trained using a 70/10/20 train/valid/test split. We train probes on 5 seeded splits with 3 random initializations per seed and a 4-option L2 regularization hyperparameter sweep (0, 1, 10, 100) per random initialization, resulting in 12 probes trained per seed. The best of the 12 probes, selected through early stopping on validation accuracy (max 800 epochs), is tested, either once on the held-out portion of the split in the all examples setting or averaged over an additional 5 held-out dataset seeds in the contrastive setting. This procedure is repeated independently for each aggregation strategy. We take our final accuracy for one experimental configuration as the averaged held-out test accuracy across the 5 seeded splits for the aggregation strategy that performed best.

\textbf{MLP and Attention Probes.} In order to keep our experiments tractable while adding a huge number of additional targets, we slightly modify the above procedure. First, we only aggregate activations with \texttt{text-nonspecial-mean}, as this strategy generally performs best in the residual setting. Next, we remove the 3 random initializations per seed, drop the number of split seeds to 3, and drop the number of held-out contrastive setting test seeds to 3. All other aspects of the experimental setup are the same. For attention probes, we report the best accuracy for each experimental configuration as the best held-out test accuracy across all of the attention head and post-attention outputs at all layers.

\textbf{Prior Harnessing Index.} In addition to reporting held-out test set probe accuracies, we calculate and report a novel metric, \textit{Prior Harnessing Index} (PHI), which represents the amount of information a model is able to glean regarding a correct final answer from a purely textual input distribution. Concretely, we define PHI as
\[
\mathrm{PHI}(Q)
=
\log p(a^* \mid Q) - \log p(a^* \mid Q_{\emptyset}).
\]
Here, \(p(a^* \mid Q)\) represents the model's probability of the correct answer \(a^*\) after conditioning on the question \(Q\), while \(p(a^* \mid Q_{\emptyset})\) represents the model's baseline probability of \(a^*\) under a null prompt (i.e., "What is the answer?").

%% file: sections/results.tex
\section{Results}
\label{sec:results}



\begin{table}[t]
\centering

\caption{\textbf{Naive Bayes classifier.} In order to determine whether or not our probes are simply picking up on surface level confounds, we train a text-based Naive Bayes classifier. The model's performance indicates that our contrastive probes pick up on deep mirage mechanisms, while all-examples probes are likely distracted by surface-level features.}
\label{tab:naive-bayes-summary}
\vspace{1mm}

\begingroup
\footnotesize
\setlength{\tabcolsep}{3pt}
\renewcommand{\arraystretch}{0.95}
\begin{tabular*}{\columnwidth}{@{\extracolsep{\fill}}lcccc@{}}
\hline
Model (Setting) & VQA-RAD & MMMU-Pro & MedXpertQA & All \\
\hline
Ovis (Contrastive) & 61.88\% & 57.67\% & \textbf{66.67\%} & 64.85\% \\
Ovis (All Examples) & 75.38\% & \textbf{81.11\%} & 67.50\% & 75.74\% \\
Qwen (Contrastive) & \textbf{59.08\%} & 54.80\% & 45.45\% & 52.24\% \\
Qwen (All Examples) & 77.08\% & \textbf{81.50\%} & 79.64\% & 72.80\% \\
\hline
\end{tabular*}
\endgroup
\vspace{1mm}
\end{table}



\subsection{Mirage Mechanisms Are Linearly Represented in Image-Present Latent Space}
\label{subsec:mirage_mechanisms_linear}

As presented in Table~\ref{tab:residual-attention-probing-side-by-side}, we demonstrate that mirage behavior is clearly represented across model layer components through non-trivial probing accuracy. When targeting the residual stream, in Ovis's contrastive and all-examples settings, we achieve linear probe accuracies of up to 75\% and 73\%, respectively. Similarly, in Qwen's contrastive and all-examples settings, we achieve linear probe accuracies of up to 72\% and 65\%. Attention head and post-attention targets show similar results, with Ovis accuracies of up to 74\% and 85\% and Qwen accuracies of up to 81\% and 69\%, respectively. Generally, across all contrastive settings, multi-layer perceptron and concatenated-activations probes do not significantly outperform logistic regression, indicating that mirage mechanisms are represented linearly in latent space. We note that difference-in-activations probes achieve remarkably high accuracy, motivating their use as mirage-behavior classifiers. Probe strategy results on MLP targets, which follow a generally similar trend, are presented in Appendix~\ref{app:mlp_targets}.

\subsection{The \textit{Mirage Probes} Framework Effectively Reduces Surface Level Confounds}
\label{subsec:mirage_probes_framework}

Because with-image mirages are invisible from a black-box perspective, we cannot use causal steering to verify that our probes have truly picked up on a mirage representation. Accordingly, we train a Naive Bayes textual classifier following a similar procedure to that used for probe training in order to determine whether or not our probes are detecting surface-level patterns or deeper mirage mechanisms. Detailed training details are discussed in Appendix~\ref{app:naive_training}, and results are presented in Table~\ref{tab:naive-bayes-summary}. Overall, we find that our contrastive \textit{Mirage Probes} framework significantly reduces surface-level separability due to its low semantic and topical variance across class examples. In particular, we note that contrastive setting accuracy ranges from 45\% to 66\% while all-examples setting accuracy ranges from 67\% to 81\%, emphasizing the reduction in confounds our framework is able to achieve. Crucially, contrastive logistic regression probe accuracy remains higher than Naive Bayes accuracy across all models, benchmarks, and targets, indicating with high confidence that our  probes pick up on deeper mechanistic features. However, we note that increases in probe accuracy on the MMMU-Pro and MedXpertQA benchmarks in the all-examples setting are directly and consistently correlated with sharp increases in surface-level separability, indicating that these probes are likely picking up on textual cues instead of identifying the desired mirage mechanism. Further analysis of per-dataset textual confounds is presented in Appendix~\ref{app:phrase_group}.

\begin{table}[t]
\centering

\caption{\textbf{Image reliance per benchmark.} We use GPT-5-mini to annotate perceived image reliance across three benchmarks: VQA-RAD, MMMU-Pro, and MedXpertQA. We find that VQA-RAD is the only dataset wherein questions that generate mirage responses are consistently classified as image-reliant, supporting a notion of a distinct, spurious-image-based mirage mechanism.}
\label{tab:mirage-like-counts-2x2}
\vspace{1mm}

\scriptsize
\resizebox{\columnwidth}{!}{%
\begin{tabular}{@{}ccc@{}}
\toprule
\textbf{VQA-RAD} & \textbf{MMMU-Pro} & \textbf{MedXpertQA} \\
\midrule

\begin{tabular}{@{}lrr@{}}
\toprule
 & + Mirage & - Mirage \\
\midrule
Image-Free & 60 & 3 \\
Image-Reliant & 2909 & 236 \\
\bottomrule
\end{tabular}
&
\begin{tabular}{@{}lrr@{}}
\toprule
 & + Mirage & - Mirage \\
\midrule
Image-Free & 836 & 18 \\
Image-Reliant & 1139 & 1421 \\
\bottomrule
\end{tabular}
&
\begin{tabular}{@{}lrr@{}}
\toprule
 & + Mirage & - Mirage \\
\midrule
Image-Free & 2399 & 15 \\
Image-Reliant & 2413 & 260 \\
\bottomrule
\end{tabular}
\\

\bottomrule
\end{tabular}%
}
\vspace{1mm}
\end{table}

\subsection{Two Distinct Mirage Mechanisms: Spurious Images and Textual Biases}
\label{subsec:two_distinct_mirage}


Across both models and targets, in the contrastive setting, we notice that VQA-RAD examples are consistently more separable than those drawn from the other two datasets. Furthermore, we note that the accuracy achieved by VQA-RAD probes is often higher than the accuracy achieved by training on all three datasets together. We attribute this higher separability to the presence of two markedly different mirage mechanisms, \textit{spurious images} and \textit{textual biases}, which are determined by different directions in latent space and thus impede separability when simultaneously present across mirage-labeled examples. We define the usage of spurious images as the use of information pulled from visual priors, or in other words cases where the model constructs false images in latent space in order to answer the user's query. Differently, we define the usage of textual biases as the use of purely textual priors to answer the provided question, with the model ignoring the presence of visual information altogether.

We hypothesize that these two distinct mechanisms are natural consequences of VLM reward hacking~\citep{skalse2025defining} during the reinforcement learning process and should thus respectively occur when they represent the path of least resistance to answering a question correctly. More concretely, we believe that textual priors will be used to answer any question wherein the model is able to determine a single answer with high confidence without any access to visual information. Spurious images will only be created when the provided textual distribution is not sufficiently rich to answer the question but is instead able to evoke false visual priors that can lead to a correct answer.

We provide evidence for this interpretation in Table~\ref{tab:mirage-like-counts-2x2} and Table~\ref{tab:phi-transfer-side-by-side}. First, in Table~\ref{tab:mirage-like-counts-2x2}, we use a few-shot prompted version of GPT-5-mini~\citep{openai2025gpt5} to annotate each benchmark’s questions for perceived image dependence, marking questions that a human could not answer without referencing the provided visual material as image-reliant, and questions that a human could answer from the text alone as image-free. We find that VQA-RAD is the only dataset among the three wherein mirage-like examples are almost always image-reliant, while in the other two datasets, questions are evenly split. This supports the hypothesis that lower probe accuracy for the other benchmarks may be caused by competing mirage directions, as it seems likely that perceived image-reliance is correlated with the richness of the provided textual distribution, and thus VQA-RAD mirages mostly utilize spurious images while the other benchmarks' mirages rely equally on both mechanisms.

However, as illustrated in Table~\ref{tab:phi-transfer-side-by-side} (right), notions of human image-reliance and VLM image-reliance are not identical. We train contrastive logistic regression probes on Qwen activations using only image-reliant and non-image reliant examples, respectively, from the MMMU-Pro and MedXpertQA benchmarks. We find that filtering training examples solely based on this criterion does not increase separability. Accordingly, we develop Prior Harnessing Index (described in Section~\ref{sec:exp_setup}) as a more accurate representation of VLM image-reliance, as it quantifies the amount of information the model itself is able to get out of the textual distribution without relying on visual cues. As presented in Table~\ref{tab:phi-transfer-side-by-side} (left), while Ovis is able to gain significant correct answer confidence from the textual distributions found in MMMU-Pro and MedXpertQA questions, VQA-RAD questions differently increase uncertainty. Furthermore, only in VQA-RAD are mirage labels highly correlated with low PHI, again explaining why mirage probe separability is higher in the VQA-RAD setting. We thus conclude that it is a combination of perceived image-reliance and the richness of the provided textual distribution that determine which mirage mechanism will be activated for a given question.

%% file: sections/conclusions.tex
\begin{table*}[t]
\centering

\caption{\textbf{VLM image-reliance is not equivalent to human image-reliance}. The table on the right shows results attained by training contrastive logistic regression probes on only human-perceived image-free or image-reliant examples from the MMMU-Pro and MedXpertQA datasets. Human-perceived image-reliance conditioning does not improve probe separation. The table on the left shows mirage correlation with PHI scores and mean PHI scores per benchmark. VQA-RAD is the only dataset wherein the question text itself does not significantly boost answer confidence and there is a strong correlation between low PHI and mirage behavior.}
\label{tab:phi-transfer-side-by-side}
\vspace{1mm}

\begin{minipage}[t]{0.44\textwidth}
\centering
\small
\setlength{\tabcolsep}{3pt}
\renewcommand{\arraystretch}{0.95}
\begin{tabular}{@{}>{\raggedright\arraybackslash}p{0.36\linewidth}
                >{\centering\arraybackslash}p{0.27\linewidth}
                >{\centering\arraybackslash}p{0.27\linewidth}@{}}
\toprule
Dataset & Mirage Corr. & Mean PHI \\
\midrule
VQA-RAD & -0.36 & -0.41 \\
MMMU-Pro & -0.23 & 1.20 \\
MedXpertQA & 0.14 & 0.30 \\
\bottomrule
\end{tabular}

\par\vspace{2.5mm}
\noindent\textbf{(a) Ovis PHI}
\end{minipage}
\hspace{0.03\textwidth}
\begin{minipage}[t]{0.50\textwidth}
\centering
\small
\setlength{\tabcolsep}{3pt}
\renewcommand{\arraystretch}{0.95}
\begin{tabular}{@{}>{\raggedright\arraybackslash}p{0.34\linewidth}
                >{\centering\arraybackslash}p{0.30\linewidth}
                >{\centering\arraybackslash}p{0.30\linewidth}@{}}
\toprule
 & Train Image-Free & Train Image-Reliant \\
\midrule
Test Image-Free & 60.00\% & 58.20\% \\
Test Image-Reliant & 61.36\% & 58.00\% \\
\bottomrule
\end{tabular}

\par\vspace{2.5mm}
\noindent\textbf{(b) Reliance-conditioned training}
\end{minipage}
\vspace{1mm}
\end{table*}

\section{Conclusion}

We introduce \emph{Mirage Probes}, a representation-level framework for studying mirage behavior in VLMs. Across two open-source VLMs and three VQA benchmarks, we find that mirage-associated generations are linearly decodable from image-present internal activations across residual stream, MLP, post-attention, and attention-head sites. Contrastive difference probes recover this signal most cleanly, while two-layer MLP probes do not meaningfully outperform linear probes, suggesting that mirage information is encoded along recoverable linear directions rather than requiring nonlinear extraction. A Naive Bayes text baseline remains weaker, indicating that the contrastive probe signal is not reducible to surface lexical features.

Our results further suggest that behavioral mirages are not a single phenomenon. Cross-benchmark separability patterns are consistent with two regimes: \emph{spurious images}, where the model behaves as though unsupported visual content were present in latent space, and \emph{textual biases}, where it answers from text-distributional priors without engaging visual representations. We view this decomposition as an empirically motivated hypothesis, not a causal claim, and validating it will require interventions such as steering, ablations, or activation patching. The mirage mechanism distinction matters for mitigation: text-distribution cleaning may reduce textual-bias mirages, but spurious-image mirages likely require representational interventions. The \emph{Mirage Probes} framework provides a starting point for diagnosing and ultimately improving faithful visual grounding in VLMs.

%% file: sections/limitations.tex
\section{Limitations and Broader Impact}
\label{sec:limitations_and_broader_impact}
Several limitations qualify our findings. Our mirage and non-mirage labels rely on heuristic criteria, which could introduce label noise. Because with-image mirages are not observable from outputs, we cannot causally validate probe-identified directions; our Naive Bayes baseline rules out a specific class of surface-feature confounds, but the central claim remains correlational. Our two-mechanism account (spurious images versus textual priors) is based on cross-benchmark separability differences and should be read as a hypothesis motivating further mechanistic work rather than a settled finding. As diagnostic interpretability research, this work does not introduce new generative capabilities. Its primary positive impact is enabling more faithful evaluation of VLMs in safety-critical domains such as medical image interpretation, where ungrounded mirage responses pose direct risks to users. We do not see a direct path to negative societal impact beyond risks already present in the underlying open-source models we probe.

\section*{Acknowledgments}
We thank the NSF AI Institute in Dynamic Systems (Grant \#2112085) for funding our work.

%% file: sections/appendix.tex
\section{Dataset Sizes}\label{app:dataset_sizes}

\begin{table}[htbp]
\centering

\caption{Post-filtering dataset sizes for contrastive and all-examples probe datasets.}
\label{tab:probe-dataset-sample-counts}
\vspace{1mm}

\small
\setlength{\tabcolsep}{5pt}
\renewcommand{\arraystretch}{0.95}
\begin{tabular}{llrr}
\toprule
Model & Benchmark & Contrastive Samples & All-Examples Samples \\
\midrule
Ovis & VQA-RAD & 98 & 235 \\
Ovis & MMMU-Pro & 48 & 144 \\
Ovis & MedXpertQA & 18 & 33 \\
Ovis & All & 164 & 406 \\
\midrule
Qwen & VQA-RAD & 222 & 418 \\
Qwen & MMMU-Pro & 314 & 792 \\
Qwen & MedXpertQA & 202 & 410 \\
Qwen & All & 738 & 1356 \\
\bottomrule
\end{tabular}
\end{table}


\section{Naive Bayes Training Details}\label{app:naive_training}

For the Naive Bayes text-confound baseline, we train a Laplace-smoothed binary multinomial model over lexical indicators extracted from model responses. Each example is represented by the set of unigrams and bigrams present in the response, using binary presence/absence features rather than counts. Class scores are computed as
\[
\log p(y) + \sum_{g \in x} \log p(g \mid y),
\]
with add-one smoothing applied to both class priors and feature likelihoods. As in probe training, examples with fewer than 10 response tokens are discarded.

For each model/setting/benchmark configuration, we run five seeds. In the all-examples setting, we first construct a class-balanced and benchmark-balanced subset, targeting 500 examples per class when available, and then use an 80/20 class-stratified train/test split. In the contrastive setting, we train on the full contrastive pool for the selected benchmark(s) and evaluate on a class-balanced, benchmark-stratified subset drawn from all-examples responses after removing any overlap with contrastive training questions.

We report mean accuracy across the five seeds. We additionally report cross-benchmark transfer by training on one benchmark and testing on another under the same balancing protocol (see below).

\section{Naive Bayes Cross-Benchmark Transfer}

\begin{table}[htbp]
\centering

\caption{Naive Bayes classifier transfer metrics.}
\label{tab:cross-benchmark-transfer}
\vspace{1mm}

\small
\resizebox{\columnwidth}{!}{%
\begin{tabular}{lcccccc}
\hline
Model (Setting) & \texttt{vqa}$\rightarrow$\texttt{mmmu} & \texttt{vqa}$\rightarrow$\texttt{medx} & \texttt{mmmu}$\rightarrow$\texttt{vqa} & \texttt{mmmu}$\rightarrow$\texttt{medx} & \texttt{medx}$\rightarrow$\texttt{vqa} & \texttt{medx}$\rightarrow$\texttt{mmmu} \\
\hline
Ovis (Contrastive) & 52.00\% & 53.33\% & 73.62\% & 16.67\% & 50.43\% & 53.83\% \\
Ovis (All-Examples) & 54.66\% & 49.47\% & 53.10\% & 50.00\% & 52.48\% & 52.05\% \\
Qwen (Contrastive) & 51.26\% & 46.87\% & 41.84\% & 49.48\% & 54.29\% & 51.36\% \\
Qwen (All-Examples) & 46.34\% & 47.45\% & 48.66\% & 51.31\% & 49.71\% & 56.92\% \\
\hline
\end{tabular}%
}
\end{table}


\clearpage
\section{Phrase-Group Confounds}\label{app:phrase_group}

\begin{table*}[htbp]
\centering

\caption{Phrase-group class separators for Ovis. Cells show mirage-like example match rate minus non-mirage-like example match rate.}
\label{tab:phrase-separators-ovis}
\vspace{1mm}

\scriptsize
\setlength{\tabcolsep}{3pt}
\renewcommand{\arraystretch}{0.92}

\begin{minipage}[t]{0.48\textwidth}
\centering
\textbf{(a) Ovis Contrastive}

\vspace{1mm}
\resizebox{\linewidth}{!}{%
\begin{tabular}{lccc}
\hline
Phrase Group & VQA-RAD & MMMU-Pro & MedXpertQA \\
\hline
reasoning\_scaffold & -8.2\% & +0.0\% & -11.1\% \\
image\_grounding & -2.0\% & +4.2\% & +11.1\% \\
uncertainty & -8.2\% & -8.3\% & -11.1\% \\
hedging & -2.0\% & -16.7\% & +0.0\% \\
answer\_boilerplate & -12.2\% & +4.2\% & +0.0\% \\
radiology\_terms & +0.0\% & +4.2\% & -11.1\% \\
\hline
\end{tabular}%
}
\end{minipage}
\hfill
\begin{minipage}[t]{0.48\textwidth}
\centering
\textbf{(b) Ovis All-Examples}

\vspace{1mm}
\resizebox{\linewidth}{!}{%
\begin{tabular}{lccc}
\hline
Phrase Group & VQA-RAD & MMMU-Pro & MedXpertQA \\
\hline
reasoning\_scaffold & -15.1\% & -32.8\% & -11.0\% \\
image\_grounding & +8.7\% & +7.1\% & +10.6\% \\
uncertainty & -22.6\% & -3.7\% & -9.8\% \\
hedging & -22.4\% & -7.6\% & -7.1\% \\
answer\_boilerplate & -6.4\% & +11.7\% & -0.1\% \\
radiology\_terms & -3.9\% & +10.2\% & -1.0\% \\
\hline
\end{tabular}%
}
\end{minipage}
\end{table*}

\begin{table*}[htbp]
\centering

\caption{Phrase-group class separators for Qwen. Cells show mirage-like example match rate minus non-mirage-like example match rate.}
\label{tab:phrase-separators-qwen}
\vspace{1mm}

\scriptsize
\setlength{\tabcolsep}{3pt}
\renewcommand{\arraystretch}{0.92}

\begin{minipage}[t]{0.48\textwidth}
\centering
\textbf{(a) Qwen Contrastive}

\vspace{1mm}
\resizebox{\linewidth}{!}{%
\begin{tabular}{lccc}
\hline
Phrase Group & VQA-RAD & MMMU-Pro & MedXpertQA \\
\hline
reasoning\_scaffold & +0.0\% & -1.9\% & +5.9\% \\
image\_grounding & +1.8\% & -0.6\% & -4.0\% \\
uncertainty & +0.9\% & +0.0\% & -1.0\% \\
hedging & +3.6\% & -4.5\% & +1.0\% \\
answer\_boilerplate & +0.9\% & +0.6\% & -5.9\% \\
radiology\_terms & +0.0\% & +2.5\% & +2.0\% \\
\hline
\end{tabular}%
}
\end{minipage}
\hfill
\begin{minipage}[t]{0.48\textwidth}
\centering
\textbf{(b) Qwen All-Examples}

\vspace{1mm}
\resizebox{\linewidth}{!}{%
\begin{tabular}{lccc}
\hline
Phrase Group & VQA-RAD & MMMU-Pro & MedXpertQA \\
\hline
reasoning\_scaffold & -4.0\% & +3.6\% & +7.8\% \\
image\_grounding & +3.0\% & -3.8\% & -32.0\% \\
uncertainty & -10.1\% & -0.1\% & +0.8\% \\
hedging & -14.3\% & +8.4\% & +4.9\% \\
answer\_boilerplate & +0.8\% & +14.7\% & -8.6\% \\
radiology\_terms & +0.0\% & +2.6\% & +0.8\% \\
\hline
\end{tabular}%
}
\end{minipage}
\end{table*}

\section{MLP Probe Results}\label{app:mlp_targets}

\begin{table*}[htbp]
\centering

\caption{Probe strategy performance on MLP-target activations.}
\label{tab:mlp-target-all-probes}
\begin{minipage}[htbp]{0.60\textwidth}
\centering
\vspace{1mm}

\begingroup
\fontsize{6.9}{7.8}\selectfont
\setlength{\tabcolsep}{0.8pt}
\renewcommand{\arraystretch}{0.86}
\setlength{\aboverulesep}{0pt}
\setlength{\belowrulesep}{2pt}
\begin{tabular*}{\linewidth}{@{\extracolsep{\fill}}lcccc@{}}
\toprule
Strategy & VQA-RAD & MMMU-Pro & MedXpertQA & All \\
\midrule

\multicolumn{5}{c}{\textbf{Ovis -- Contrastive}} \\
\midrule
LogReg & \textbf{76.33\%} & 61.67\% & 63.33\% & 72.00\% \\
MLP    & \textbf{71.33\%} & 60.00\% & 66.67\% & 69.33\% \\
Concat & 68.33\% & 49.00\% & 53.33\% & \textbf{70.67\%} \\
Diff   & 88.33\% & \textbf{91.33\%} & 58.33\% & 89.00\% \\

\midrule
\multicolumn{5}{c}{\textbf{Ovis -- All Examples}} \\
\midrule
LogReg & \textbf{71.63\%} & 67.82\% & 61.90\% & 69.96\% \\
MLP    & 71.63\% & 65.52\% & \textbf{85.71\%} & 70.37\% \\
Concat & 63.83\% & \textbf{66.67\%} & 28.57\% & 60.49\% \\
Diff   & \textbf{95.04\%} & 86.21\% & 71.43\% & 90.95\% \\

\midrule
\multicolumn{5}{c}{\textbf{Qwen -- Contrastive}} \\
\midrule
LogReg & \textbf{74.00\%} & 59.67\% & 60.00\% & 59.67\% \\
MLP    & \textbf{70.33\%} & 60.33\% & 59.67\% & 62.33\% \\
Concat & \textbf{62.67\%} & 56.67\% & 47.33\% & 57.00\% \\
Diff   & \textbf{93.33\%} & 78.00\% & 80.00\% & 81.67\% \\

\midrule
\multicolumn{5}{c}{\textbf{Qwen -- All Examples}} \\
\midrule
LogReg & 59.13\% & 63.08\% & \textbf{64.23\%} & 56.99\% \\
MLP    & 71.36\% & \textbf{75.65\%} & 68.60\% & 72.91\% \\
Concat & 67.14\% & \textbf{71.88\%} & 64.25\% & 69.41\% \\
Diff   & \textbf{95.77\%} & 80.00\% & 76.81\% & 84.93\% \\

\bottomrule
\end{tabular*}
\endgroup
\end{minipage}
\end{table*}


\section{Compute Resources}
All experiments are run using one or more NVIDIA RTX A6000s, each with 50 gigabytes of VRAM. Full Ovis runs on residual or additional targets (MLP, attention heads, post-attention) take around 12 hours to run with tasks orchestrated across two GPUs. Qwen runs require around 24 hours of activation extraction across four GPUs followed by 12 hours of probing orchestrated across two. All experiments (except for the Qwen activation extraction) could plausibly be run with much less powerful GPUs.

%% file: neurips_2026.bib
@inproceedings{yue2024mmmu,
  title={Mmmu: A massive multi-discipline multimodal understanding and reasoning benchmark for expert agi},
  author={Yue, Xiang and Ni, Yuansheng and Zhang, Kai and Zheng, Tianyu and Liu, Ruoqi and Zhang, Ge and Stevens, Samuel and Jiang, Dongfu and Ren, Weiming and Sun, Yuxuan and others},
  booktitle={Proceedings of the IEEE/CVF conference on computer vision and pattern recognition},
  pages={9556--9567},
  year={2024}
}

@article{lau2018dataset,
  title={A dataset of clinically generated visual questions and answers about radiology images},
  author={Lau, Jason J and Gayen, Soumya and Ben Abacha, Asma and Demner-Fushman, Dina},
  journal={Scientific data},
  volume={5},
  number={1},
  pages={180251},
  year={2018},
  publisher={Nature Publishing Group}
}

@inproceedings{li2022blip,
  title={Blip: Bootstrapping language-image pre-training for unified vision-language understanding and generation},
  author={Li, Junnan and Li, Dongxu and Xiong, Caiming and Hoi, Steven},
  booktitle={International conference on machine learning},
  pages={12888--12900},
  year={2022},
  organization={PMLR}
}

@article{liu2023visual,
  title={Visual instruction tuning},
  author={Liu, Haotian and Li, Chunyuan and Wu, Qingyang and Lee, Yong Jae},
  journal={Advances in neural information processing systems},
  volume={36},
  pages={34892--34916},
  year={2023}
}

@article{li2023llava,
  title={Llava-med: Training a large language-and-vision assistant for biomedicine in one day},
  author={Li, Chunyuan and Wong, Cliff and Zhang, Sheng and Usuyama, Naoto and Liu, Haotian and Yang, Jianwei and Naumann, Tristan and Poon, Hoifung and Gao, Jianfeng},
  journal={Advances in Neural Information Processing Systems},
  volume={36},
  pages={28541--28564},
  year={2023}
}

@inproceedings{moor2023med,
  title={Med-flamingo: a multimodal medical few-shot learner},
  author={Moor, Michael and Huang, Qian and Wu, Shirley and Yasunaga, Michihiro and Dalmia, Yash and Leskovec, Jure and Zakka, Cyril and Reis, Eduardo Pontes and Rajpurkar, Pranav},
  booktitle={Machine learning for health (ML4H)},
  pages={353--367},
  year={2023},
  organization={PMLR}
}

@article{asadi2026mirage,
  title={Mirage: The illusion of visual understanding},
  author={Asadi, Mohammad and O'Sullivan, Jack W and Cao, Fang and Nedaee, Tahoura and Rajabalifardi, Kamyar and Li, Fei-Fei and Adeli, Ehsan and Ashley, Euan},
  journal={arXiv preprint arXiv:2603.21687},
  year={2026}
}

@inproceedings{li2023evaluating,
  title={Evaluating object hallucination in large vision-language models},
  author={Li, Yifan and Du, Yifan and Zhou, Kun and Wang, Jinpeng and Zhao, Xin and Wen, Ji-Rong},
  booktitle={Proceedings of the 2023 conference on empirical methods in natural language processing},
  pages={292--305},
  year={2023}
}

@inproceedings{antol2015vqa,
  title={Vqa: Visual question answering},
  author={Antol, Stanislaw and Agrawal, Aishwarya and Lu, Jiasen and Mitchell, Margaret and Batra, Dhruv and Zitnick, C Lawrence and Parikh, Devi},
  booktitle={Proceedings of the IEEE international conference on computer vision},
  pages={2425--2433},
  year={2015}
}

@inproceedings{mathew2021docvqa,
  title={Docvqa: A dataset for vqa on document images},
  author={Mathew, Minesh and Karatzas, Dimosthenis and Jawahar, CV},
  booktitle={Proceedings of the IEEE/CVF winter conference on applications of computer vision},
  pages={2200--2209},
  year={2021}
}

@article{kafle2017visual,
  title={Visual question answering: Datasets, algorithms, and future challenges},
  author={Kafle, Kushal and Kanan, Christopher},
  journal={Computer Vision and Image Understanding},
  volume={163},
  pages={3--20},
  year={2017},
  publisher={Elsevier}
}

@inproceedings{kim2022ocr,
  title={Ocr-free document understanding transformer},
  author={Kim, Geewook and Hong, Teakgyu and Yim, Moonbin and Nam, JeongYeon and Park, Jinyoung and Yim, Jinyeong and Hwang, Wonseok and Yun, Sangdoo and Han, Dongyoon and Park, Seunghyun},
  booktitle={European Conference on Computer Vision},
  pages={498--517},
  year={2022},
  organization={Springer}
}

@article{lu2022learn,
  title={Learn to explain: Multimodal reasoning via thought chains for science question answering},
  author={Lu, Pan and Mishra, Swaroop and Xia, Tanglin and Qiu, Liang and Chang, Kai-Wei and Zhu, Song-Chun and Tafjord, Oyvind and Clark, Peter and Kalyan, Ashwin},
  journal={Advances in neural information processing systems},
  volume={35},
  pages={2507--2521},
  year={2022}
}

@article{zhang2023pmc,
  title={Pmc-vqa: Visual instruction tuning for medical visual question answering},
  author={Zhang, Xiaoman and Wu, Chaoyi and Zhao, Ziheng and Lin, Weixiong and Zhang, Ya and Wang, Yanfeng and Xie, Weidi},
  journal={arXiv preprint arXiv:2305.10415},
  year={2023}
}

@article{yue2025mmmupro,
  title={Mmmu-pro: A more robust multi-discipline multimodal understanding benchmark},
  author={Yue, Xiang and Zheng, Tianyu and Ni, Yuansheng and Wang, Yubo and Zhang, Kai and Tong, Shengbang and Sun, Yuxuan and Yu, Botao and Zhang, Ge and Sun, Huan and others},
  booktitle={Proceedings of the 63rd Annual Meeting of the Association for Computational Linguistics (Volume 1: Long Papers)},
  pages={15134--15186},
  year={2025}
}

@article{zuo2025medxpertqa,
  title={Medxpertqa: Benchmarking expert-level medical reasoning and understanding},
  author={Zuo, Yuxin and Qu, Shang and Li, Yifei and Chen, Zhangren and Zhu, Xuekai and Hua, Ermo and Zhang, Kaiyan and Ding, Ning and Zhou, Bowen},
  journal={arXiv preprint arXiv:2501.18362},
  year={2025}
}

@inproceedings{burgess2025microvqa,
  title={Microvqa: A multimodal reasoning benchmark for microscopy-based scientific research},
  author={Burgess, James and Nirschl, Jeffrey J and Bravo-S{\'a}nchez, Laura and Lozano, Alejandro and Gupte, Sanket Rajan and Galaz-Montoya, Jesus G and Zhang, Yuhui and Su, Yuchang and Bhowmik, Disha and Coman, Zachary and others},
  booktitle={Proceedings of the IEEE/CVF Conference on Computer Vision and Pattern Recognition},
  pages={19552--19564},
  year={2025}
}

@article{lu2024ovis,
  title={Ovis: Structural embedding alignment for multimodal large language model},
  author={Lu, Shiyin and Li, Yang and Chen, Qing-Guo and Xu, Zhao and Luo, Weihua and Zhang, Kaifu and Ye, Han-Jia},
  journal={arXiv preprint arXiv:2405.20797},
  year={2024}
}

@misc{glm2024chatglmfamilylargelanguage,
  title={ChatGLM: A Family of Large Language Models from GLM-130B to GLM-4 All Tools}, 
  author={{Team GLM} and Aohan Zeng and Bin Xu and Bowen Wang and Chenhui Zhang and Da Yin and Dan Zhang and Diego Rojas and Guanyu Feng and Hanlin Zhao and Hanyu Lai and Hao Yu and Hongning Wang and Jiadai Sun and Jiajie Zhang and Jiale Cheng and Jiayi Gui and Jie Tang and Jing Zhang and Jingyu Sun and Juanzi Li and Lei Zhao and Lindong Wu and Lucen Zhong and Mingdao Liu and Minlie Huang and Peng Zhang and Qinkai Zheng and Rui Lu and Shuaiqi Duan and Shudan Zhang and Shulin Cao and Shuxun Yang and Weng Lam Tam and Wenyi Zhao and Xiao Liu and Xiao Xia and Xiaohan Zhang and Xiaotao Gu and Xin Lv and Xinghan Liu and Xinyi Liu and Xinyue Yang and Xixuan Song and Xunkai Zhang and Yifan An and Yifan Xu and Yilin Niu and Yuantao Yang and Yueyan Li and Yushi Bai and Yuxiao Dong and Zehan Qi and Zhaoyu Wang and Zhen Yang and Zhengxiao Du and Zhenyu Hou and Zihan Wang},
  year={2024},
  eprint={2406.12793},
  archivePrefix={arXiv},
  primaryClass={cs.CL},
  url={https://arxiv.org/abs/2406.12793}, 
}

@misc{bai2025qwen3vltechnicalreport,
  title={Qwen3-VL Technical Report}, 
  author={Shuai Bai and Yuxuan Cai and Ruizhe Chen and Keqin Chen and Xionghui Chen and Zesen Cheng and Lianghao Deng and Wei Ding and Chang Gao and Chunjiang Ge and Wenbin Ge and Zhifang Guo and Qidong Huang and Jie Huang and Fei Huang and Binyuan Hui and Shutong Jiang and Zhaohai Li and Mingsheng Li and Mei Li and Kaixin Li and Zicheng Lin and Junyang Lin and Xuejing Liu and Jiawei Liu and Chenglong Liu and Yang Liu and Dayiheng Liu and Shixuan Liu and Dunjie Lu and Ruilin Luo and Chenxu Lv and Rui Men and Lingchen Meng and Xuancheng Ren and Xingzhang Ren and Sibo Song and Yuchong Sun and Jun Tang and Jianhong Tu and Jianqiang Wan and Peng Wang and Pengfei Wang and Qiuyue Wang and Yuxuan Wang and Tianbao Xie and Yiheng Xu and Haiyang Xu and Jin Xu and Zhibo Yang and Mingkun Yang and Jianxin Yang and An Yang and Bowen Yu and Fei Zhang and Hang Zhang and Xi Zhang and Bo Zheng and Humen Zhong and Jingren Zhou and Fan Zhou and Jing Zhou and Yuanzhi Zhu and Ke Zhu},
  year={2025},
  eprint={2511.21631},
  archivePrefix={arXiv},
  primaryClass={cs.CV},
  url={https://arxiv.org/abs/2511.21631}, 
}

@article{openai2024gpt4ocard,
  title={Gpt-4o system card},
  author={Hurst, Aaron and Lerer, Adam and Goucher, Adam P and Perelman, Adam and Ramesh, Aditya and Clark, Aidan and Ostrow, AJ and Welihinda, Akila and Hayes, Alan and Radford, Alec and others},
  journal={arXiv preprint arXiv:2410.21276},
  year={2024}
}

@article{alain2016understanding,
  title={Understanding intermediate layers using linear classifier probes},
  author={Alain, Guillaume and Bengio, Yoshua},
  journal={arXiv preprint arXiv:1610.01644},
  year={2016}
}

@article{belinkov2022probing,
  title={Probing classifiers: Promises, shortcomings, and advances},
  author={Belinkov, Yonatan},
  journal={Computational Linguistics},
  volume={48},
  number={1},
  pages={207--219},
  year={2022}
}

@misc{marks2024geometry,
  title={The Geometry of Truth: Emergent Linear Structure in Large Language Model Representations of True/False Datasets}, 
  author={Samuel Marks and Max Tegmark},
  year={2024},
  eprint={2310.06824},
  archivePrefix={arXiv},
  primaryClass={cs.AI},
  url={https://arxiv.org/abs/2310.06824}, 
}

@misc{burns2024discoveringlatentknowledgelanguage,
  title={Discovering Latent Knowledge in Language Models Without Supervision}, 
  author={Collin Burns and Haotian Ye and Dan Klein and Jacob Steinhardt},
  year={2024},
  eprint={2212.03827},
  archivePrefix={arXiv},
  primaryClass={cs.CL},
  url={https://arxiv.org/abs/2212.03827}, 
}

@misc{skalse2025defining,
  title={Defining and Characterizing Reward Hacking}, 
  author={Joar Skalse and Nikolaus H. R. Howe and Dmitrii Krasheninnikov and David Krueger},
  year={2025},
  eprint={2209.13085},
  archivePrefix={arXiv},
  primaryClass={cs.LG},
  url={https://arxiv.org/abs/2209.13085}, 
}

@article{openai2025gpt5,
  title={Openai gpt-5 system card},
  author={Singh, Aaditya and Fry, Adam and Perelman, Adam and Tart, Adam and Ganesh, Adi and El-Kishky, Ahmed and McLaughlin, Aidan and Low, Aiden and Ostrow, AJ and Ananthram, Akhila and others},
  journal={arXiv preprint arXiv:2601.03267},
  year={2025}
}

@inproceedings{rohrbach2018object,
  title={Object hallucination in image captioning},
  author={Rohrbach, Anna and Hendricks, Lisa Anne and Burns, Kaylee and Darrell, Trevor and Saenko, Kate},
  booktitle={Proceedings of the 2018 Conference on Empirical Methods in Natural Language Processing},
  pages={4035--4045},
  year={2018}
}

@inproceedings{sun2024aligning,
  title={Aligning large multimodal models with factually augmented rlhf},
  author={Sun, Zhiqing and Shen, Sheng and Cao, Shengcao and Liu, Haotian and Li, Chunyuan and Shen, Yikang and Gan, Chuang and Gui, Liangyan and Wang, Yu-Xiong and Yang, Yiming and others},
  booktitle={Findings of the Association for Computational Linguistics: ACL 2024},
  pages={13088--13110},
  year={2024}
}

@inproceedings{jiang2024hallucination,
  title={Hallucination augmented contrastive learning for multimodal large language model},
  author={Jiang, Chaoya and Xu, Haiyang and Dong, Mengfan and Chen, Jiaxing and Ye, Wei and Yan, Ming and Ye, Qinghao and Zhang, Ji and Huang, Fei and Zhang, Shikun},
  booktitle={Proceedings of the IEEE/CVF Conference on Computer Vision and Pattern Recognition},
  pages={27036--27046},
  year={2024}
}

@article{liu2024survey,
  title={A survey on hallucination in large vision-language models},
  author={Liu, Hanchao and Xue, Wenyuan and Chen, Yifei and Chen, Dapeng and Zhao, Xiutian and Wang, Ke and Hou, Liping and Li, Rongjun and Peng, Wei},
  journal={arXiv preprint arXiv:2402.00253},
  year={2024}
}

@article{bai2024hallucination,
  title={Hallucination of multimodal large language models: A survey},
  author={Bai, Zechen and Wang, Pichao and Xiao, Tianjun and He, Tong and Han, Zongbo and Zhang, Zheng and Shou, Mike Zheng},
  journal={arXiv preprint arXiv:2404.18930},
  year={2024}
}

@inproceedings{tong2024eyes,
  title={Eyes wide shut? exploring the visual shortcomings of multimodal llms},
  author={Tong, Shengbang and Liu, Zhuang and Zhai, Yuexiang and Ma, Yi and LeCun, Yann and Xie, Saining},
  booktitle={Proceedings of the IEEE/CVF conference on computer vision and pattern recognition},
  pages={9568--9578},
  year={2024}
}

@inproceedings{hewitt2019structural,
  title={A structural probe for finding syntax in word representations},
  author={Hewitt, John and Manning, Christopher D},
  booktitle={Proceedings of the 2019 Conference of the North American Chapter of the Association for Computational Linguistics: Human Language Technologies, Volume 1 (Long and Short Papers)},
  pages={4129--4138},
  year={2019}
}

@inproceedings{tenney2019bert,
  title={BERT rediscovers the classical NLP pipeline},
  author={Tenney, Ian and Das, Dipanjan and Pavlick, Ellie},
  booktitle={Proceedings of the 57th annual meeting of the association for computational linguistics},
  pages={4593--4601},
  year={2019}
}

@article{park2023linear,
  title={The linear representation hypothesis and the geometry of large language models},
  author={Park, Kiho and Choe, Yo Joong and Veitch, Victor},
  journal={arXiv preprint arXiv:2311.03658},
  year={2023}
}

@article{lomasov2025exploring,
  title={Exploring Human-AI Conceptual Alignment through the Prism of Chess},
  author={Lomasov, Semyon and Goldfeder, Judah and Erol, Mehmet Hamza and So, Matthew and Yan, Yao and Howard, Addison and Kutz, Nathan and Ziv, Ravid Shwartz},
  journal={arXiv preprint arXiv:2510.26025},
  year={2025}
}

@article{goldfeder2026ai,
  title={Ai must embrace specialization via superhuman adaptable intelligence},
  author={Goldfeder, Judah and Wyder, Philippe and LeCun, Yann and Ziv, Ravid Shwartz},
  journal={arXiv preprint arXiv:2602.23643},
  year={2026}
}

@misc{theodoridis2026probing,
  title={Probing Visual Concepts in Lightweight Vision-Language Models for Automated Driving}, 
  author={Nikos Theodoridis and Reenu Mohandas and Ganesh Sistu and Anthony Scanlan and Ciarán Eising and Tim Brophy},
  year={2026},
  eprint={2603.06054},
  archivePrefix={arXiv},
  primaryClass={cs.CV},
  url={https://arxiv.org/abs/2603.06054}, 
}

@article{goldowsky2025detecting,
  title={Detecting strategic deception using linear probes},
  author={Goldowsky-Dill, Nicholas and Chughtai, Bilal and Heimersheim, Stefan and Hobbhahn, Marius},
  journal={arXiv preprint arXiv:2502.03407},
  year={2025}
}
